\documentclass[twocolumn]{article}
\usepackage[utf8]{inputenc}
\usepackage{graphicx}
\usepackage{amsmath}
\usepackage{amssymb}
\usepackage{hyperref}
\usepackage{geometry}
\usepackage{float}
\usepackage{caption}

\graphicspath{ {./asset-engram/} }

\title{Adaptive Engram Memory System for Indonesian Language Model: Generative AI Based on TOBA LM for Batak and Minang Language}
\author{%
  Hokky Situngkir\footnotemark[1]
  \and
  Kevin Siringoringo\footnotemark[2]
  \and
  Andhika Bernard Lumbantobing\footnotemark[3]
}
\date{}

\begin{document}

\twocolumn[
    \maketitle
    \vspace{-0.2cm}
    \begin{center}
      \parbox{0.9\textwidth}{
        \small
        \begin{abstract}
This study presents TOBA-LM, a trilingual language model based on GPT-2 architecture with 1.2 billion parameters, trained on a corpus encompassing Indonesian, Batak, and Minangkabau using syllabic-agglutinative tokenization. The architecture integrates an \textit{Engram Memory} mechanism, an adaptive $n$-gram-based memory system with a $500{,}000 \times 768$ embedding table that captures morphological dependencies through bigram and trigram pathways. Empirical results demonstrate a training efficiency of 80\%, with the loss value dropping from 6.4 to 1.7996 in only 12,973 steps---significantly faster than the conventional transformer architecture, which required over 70,000 steps to achieve comparable convergence. These findings confirm that the integration of external statistical memory substantially reduces computational requirements for developing regional language models under limited resources.

\end{abstract}
      }
    \end{center}
    \vspace{1cm}
]

{
  \renewcommand{\thefootnote}{\fnsymbol{footnote}}
  \footnotetext[1]{AI Research Center IT Del \& Bandung Fe Institute, \texttt{hokky.situngkir@del.ac.id}}
  \footnotetext[2]{Bandung Fe Institute, \texttt{kevin@compsoc.bandungfe.net}}
  \footnotetext[3]{Bandung Fe Institute \& Adjunct Science Fellow in InaAI, \texttt{nad@compsoc.bandungfe.net}}
}

\section{Introduction}
Batak and Minang are two major languages spoken by two ethnic groups on the island of Sumatra, Indonesia. The Indonesian Statistics Agency in 2011 recorded that there are approximately 8.47 million speakers of Batak and 4.2 million speakers of Minang. The development of Large Language Models (LLMs) for regional languages in Indonesia, particularly Batak and Minangkabau, faces significant challenges due to the limited availability of high-quality datasets (low-resource languages) [1]. One of the primary obstacles in training regional language models is the use of conventional subword tokenization methods such as Byte Pair Encoding (BPE), which tends to segment words into sub-word units lacking relevant linguistic meaning, especially for languages with agglutinative characteristics. To address this issue, this study adopts the syllabic agglutinative tokenization approach pioneered by Situngkir et al. [2], who demonstrated that syllable-based token units are more effective in representing the linguistic structure of Austronesian languages compared to standard subword methods. This approach enables the preservation of linguistic information at the word-formation level, making it more suitable for regional languages rich in affixation and morphological variation [3].

The Toba Trilingual Model is built on a GPT-2 architecture utilizing syllable-based tokens with 1.2B parameters, and is trained on a trilingual corpus encompassing Indonesian, Batak, and Minangkabau. The model combines the Transformer architecture [4][5] with the Engram mechanism, a memory architecture introduced by DeepSeek to simulate retrieval mechanisms over n-gram representations that have been learned and stored within embedding weights [6]. The Engram mechanism extends the representational context of hidden states through selective n-gram integration using conditional gating, which serves to reduce noise caused by hash collisions [7] as well as linguistic phenomena such as polysemy. In the implementation of syllable-based token LLMs, this layer is designed to localize the processing of critical local dependencies between token sequences during the training phase—such as word formation and morphological rules—into a dedicated layer. The self-attention and multilayer perceptron (MLP) layers can then be allocated to modeling more complex dependencies, such as long-range dependencies and compositional reasoning.

The primary focus of this study is a review of the model training methodology combining the Transformer with the Engram, the effects of which were observed to be significant during the early pre-training phase. Through empirical observation of training logs, a sharp decline in loss values was recorded, dropping from 6.4 to 1.7966. This decline serves as an indicator of the successful integration of an appropriate tokenization system and early-sprint optimization techniques facilitated by the Engram layer, which ultimately carries substantial implications for time efficiency and computational resource utilization.

Observation of changes in the Engram memory weights indicates that the loss convergence is not merely a reduction in error values, but rather constitutes a phase transition [8][9]. As described in the Transformer Circuits study [10], this transition marks the point at which the model ceases to rely on simple frequency statistics and begins to form induction mechanisms that enable deep contextual understanding.

This study describes the hybridization of the GPT-2 architecture integrated with an Engram-based memory system. This approach offers training efficiency comparable to sparse models (MoE) while maintaining the stability of dense models. The effectiveness of this integration is confirmed through the observation of loss convergence phenomena during the early training phase (early pre-training).

\begin{figure*}[ht]
    \centering
    \includegraphics[width=\textwidth]{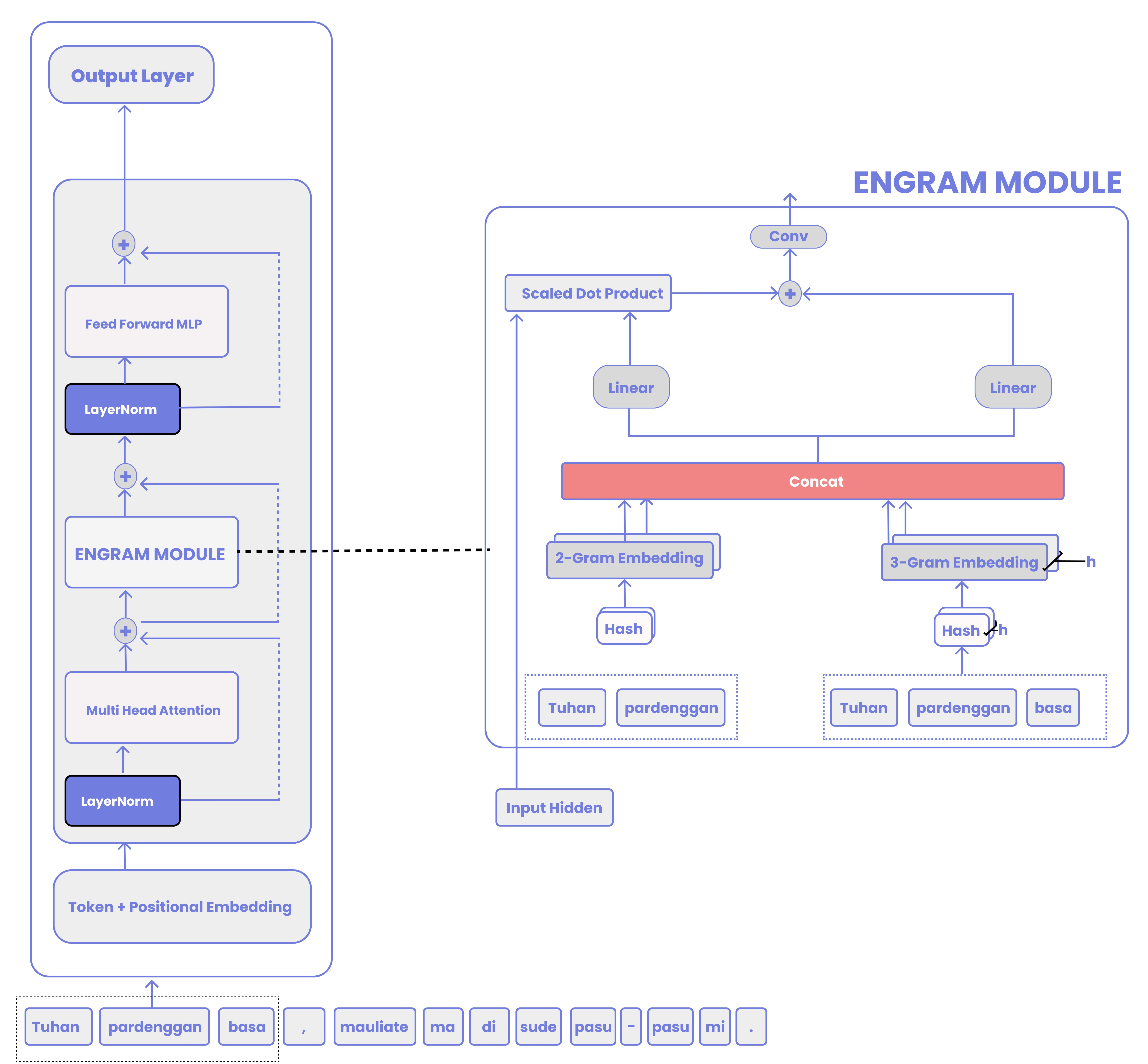}
    \caption{The architecture of TOBA-LM trilingual engram}
    \label{fig:figure1}
\end{figure*}

\section{Model Architecture and Memory System}
The implementation in this study combines a standard transformer structure with a dynamic memory system to accommodate the complex requirements of regional language data. By embedding an Engram-based memory mechanism, the model can map linguistic patterns more specifically and deeply compared to conventional architectures. This strategy underlies the loss convergence phenomenon that triggers a phase transition (phase change) at the onset of training, during which the model aggressively optimizes parameters to comprehend agglutinative language structures from the very first steps.

Figure \ref{fig:figure1} illustrates the functional integration scheme between the core transformer components and the adaptive memory module. The visualization demonstrates how hidden state representations flow through attention blocks before being further processed by the gating mechanism in the Engram memory, adapting techniques from DeepSeek [6]. This division of roles ensures that the model can simultaneously extract contextual information while retrieving statistical memory traces of regional languages.

\subsection{Transformer Decoder Architecture Configuration}
This study adopts a decoder-only transformer architecture with a 1.2B parameter configuration. The use of a billion-scale parameter model aims to enhance vector space representation of linguistic features in Indonesian, Batak, and Minang, which exhibit high morphological complexity [2]. The primary operation of the model focuses on minimizing the negative log-likelihood loss function ($L$) to predict the probability of the next token occurrence in a sequential order [11].

\begin{equation}
    L(\theta) = - \sum_{t} \log P(x_t | x_{<t}; \theta)
\end{equation}

The symbol $\theta$ represents the set of model parameters. This architecture employs Multi-Head Attention (MHA) to capture contextual dependencies between tokens. For each head $h$, the attention operation is defined as:

\begin{equation}
    \text{Attention}(Q, K, V) = \text{softmax}\left(\frac{QK^T}{\sqrt{d_k}}\right)V
\end{equation}

The massive parameter configuration provides an exceptionally stable dense foundation, relying on the principle of Scaled Dot-Product Attention to map semantic relationships between tokens within text sequences. This process begins with the linear transformation of input embeddings into three principal matrices: Q (Query), K (Key) and V (Value). The $QK^T$ multiplication operation extracts relevance scores indicating the degree of attention a given token should assign to other tokens within the sentence context. The use of $\sqrt{d_k}$ serves as a scaling factor that is crucial for maintaining gradient stability during the optimization process [12].

\subsection{Engram Memory Mechanism}
The Engram Memory mechanism is a core architectural component designed to optimize language processing for languages with high morphological density through the integration of statistical memory units (n-grams) into the transformer information pathway.

The final representation at this layer is produced through a hybrid fusion between the base representation (base embedding) and modification signals from the Engram memory units:

\begin{equation}
    h' = h + \text{Engram}(h)
\end{equation}

The memory signal is extracted through a projection function over sequential token windows stored in Memory. Based on the input flow design of the Engram module, this extraction is divided into two parallel pathways:

\begin{equation}
    E_{2gram} = \text{Lookup}(x_{t-1}, x_t)
\end{equation}

\begin{equation}
    E_{3gram} = \text{Lookup}(x_{t-2}, x_{t-1}, x_t)
\end{equation}

The variable descriptions indicate that $E_{2gram}$ represents the 2-gram embedding pathway for capturing morpheme structures and/or word formations, while $E_{3gram}$ represents the 3-gram embedding pathway for capturing broader morphophonological dependencies.

The relevance between the current hidden state and memory information is evaluated using a scaled dot-product block:

\begin{equation}
    \text{Score} = \frac{h \cdot E^T}{\sqrt{d}}
\end{equation}

The similarity score serves as a statistical anchor that contracts the search space in subsequent token prediction. The final contribution of each memory domain is regulated by an adaptive sparse gate:

\begin{equation}
    g = \sigma(W_g h)
\end{equation}

The weight matrix $W_g$ along with the control parameters ensure that only memory information with high semantic relevance is selectively activated. The implementation of this series of formulations has been empirically demonstrated to trigger aggressive convergence during the early training phase.

\section{Trilingual Corpus and Data Cleaning}
The quality and diversity of the data corpus constitute determining factors in the successful internalization of linguistic dependencies. This study compiled a comprehensive dataset to ensure that the Toba Trilingual model captures the morphology of Indonesian as well as the regional languages of Batak and Minangkabau. The constructed trilingual corpus integrates Indonesian Wikipedia, Batak Wikipedia, and Minang Wikipedia. Data enrichment was carried out through the utilization of the NusaX dataset [13], cultural literature from the Perpustakaan Digital Budaya Indonesia (PDBI; Indonesian Digital Cultural Library) [14], as well as Batak-language book corpora and song lyrics to capture language register variations ranging from formal to colloquial. The integration of the FineWeb dataset and machine translation outputs was also applied to strengthen cross-lingual semantic comprehension and data availability for the model. The unification of these diverse language domains aims to create a linguistic synergy that triggers an early phase transition within the model's internal layers. Information integrity within the trilingual corpus is ensured through a systematic data cleaning operational pipeline.

\begin{figure*}[ht]
    \centering
    \includegraphics[width=\textwidth]{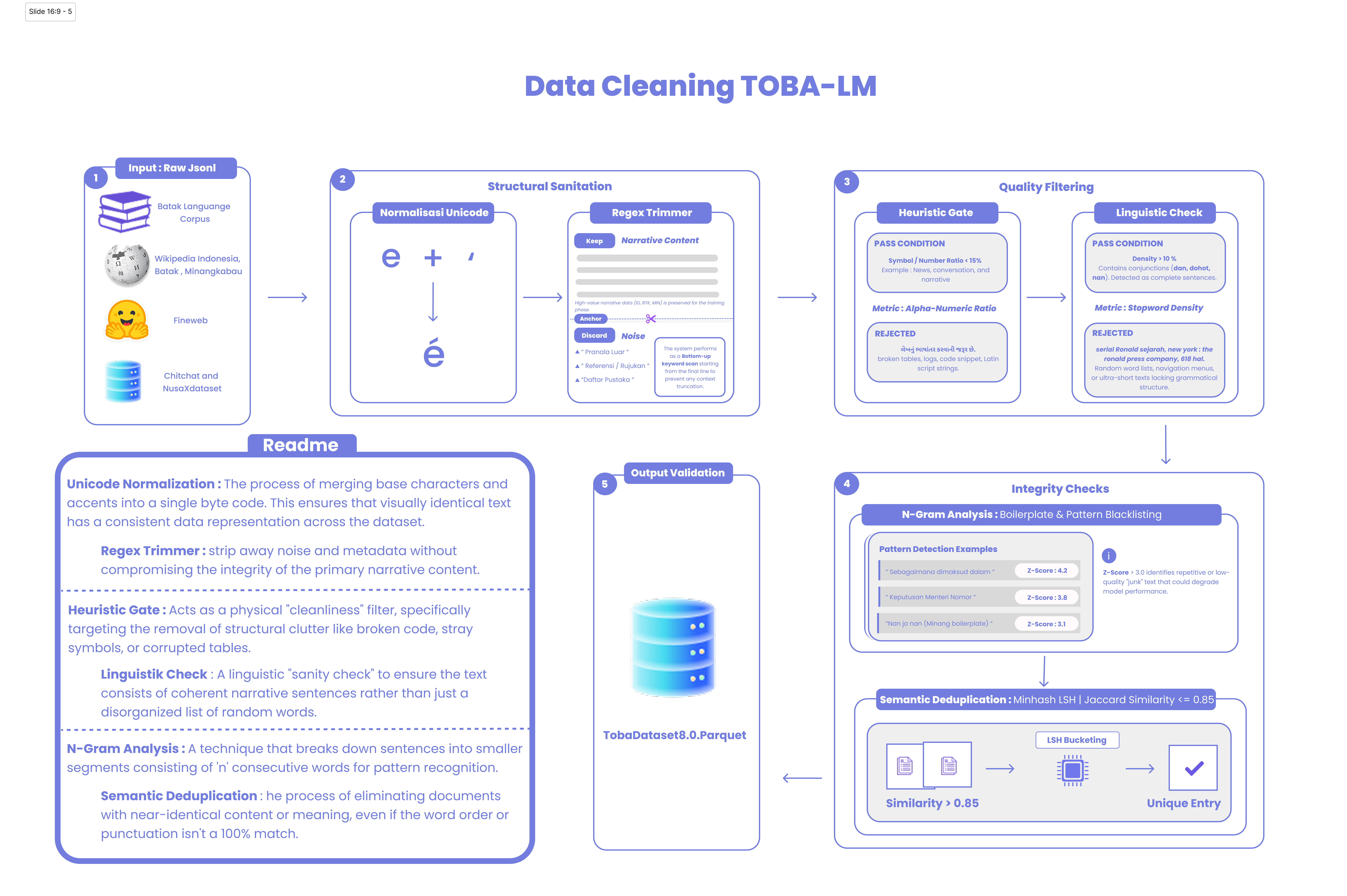}
    \caption{Toba-LM Trilingual Data Cleaning}
    \label{fig:figure2}
\end{figure*}

The series of data cleaning procedures collectively minimizes the model's workload in recognizing surface-level text patterns to ensure semantic integrity and corpus quality by integrating four main pillars of data standardization:
\begin{itemize}
    \item \textbf{Structural Sanitization.} Unicode normalization to unify base characters and accents into single, consistent byte codes, along with the use of a Regex Trimmer to eliminate noise and metadata without compromising the primary narrative content.
    \item \textbf{Quality Filtration.} Content filtering through a Heuristic Gate that removes code fragments, symbols, or corrupted tables, followed by a linguistic check to ensure natural and narrative-quality language.
    \item \textbf{Integrity Verification.} In-depth testing is applied using n-gram analysis to blacklist boilerplate patterns, as well as semantic deduplication based on MinHash LSH with a Jaccard Similarity threshold of $> 0.85$.
    \item \textbf{Final Validation and Storage.} Data that has met the quality standards is then validated and stored in Parquet format.
\end{itemize}

\section{Model Training}

\subsection{Technical Specifications and Experiment Configuration}
The detailed architecture configuration and experiment parameters used in the development of Toba Trilingual Engram are presented systematically to provide a comprehensive technical overview.

\begin{table*}[ht]
\centering
\caption{Architecture Configuration and Parameters}
\begin{tabular}{|l|l|l|}
\hline
\textbf{Category} & \textbf{Main Parameter} & \textbf{Technical Specification} \\ \hline
Backbone Architecture & Layers, Dim, Heads & 36 Blocks, d = 1280, 20 Heads \\ \hline
 & Context Length & 1,024 tokens \\ \hline
Engram Module & Table \& Embed Dim & 500,000 $\times$ 768 \\ \hline
 & Internal Heads & 8 Heads \\ \hline
Tokenization & Type & Tokenizer \\ \hline
 & Vocab Size & 2,843 Units \\ \hline
Optimization & Batch Size & 36 $\times$ 4 (effective 128) \\ \hline
 & Format & Bfloat16 (BF16) \\ \hline
\end{tabular}
\end{table*}

\subsection{Empirical Evaluation of Convergence Rate}
\begin{figure*}[ht]
    \centering
    \includegraphics[width=\textwidth]{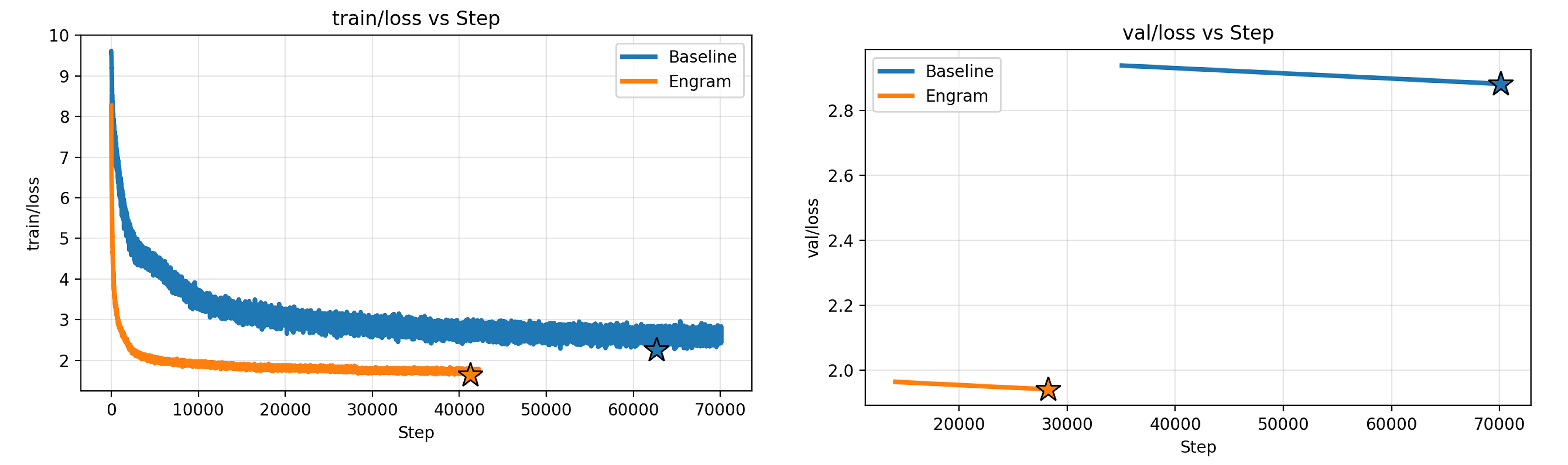}
    \caption{Comparison of TOBA-LM Trilingual Engram vs. Baseline.}
    \label{fig:figure3}
\end{figure*}

The evaluation of the Engram mechanism's effectiveness was conducted through observation of the optimization trajectory of the 1.2B parameter model. The visualization in Figure \ref{fig:figure3} underscores the effectiveness of the Engram strategy in significantly reducing computational overhead.
The training loss curve exhibits a highly aggressive decline under the Engram scenario compared to the conventional transformer baseline. The model successfully reached a loss value of 1.7996 in only 12,973 steps, while the baseline scenario remained stalled at a considerably higher loss value even after exceeding 70,000 steps. This phenomenon demonstrates that the integration of syllable-based statistical memory enables the model to internalize local linguistic patterns instantaneously without requiring the global attention process to reach convergence.

\begin{figure*}[ht]
    \centering
    \includegraphics[width=\textwidth]{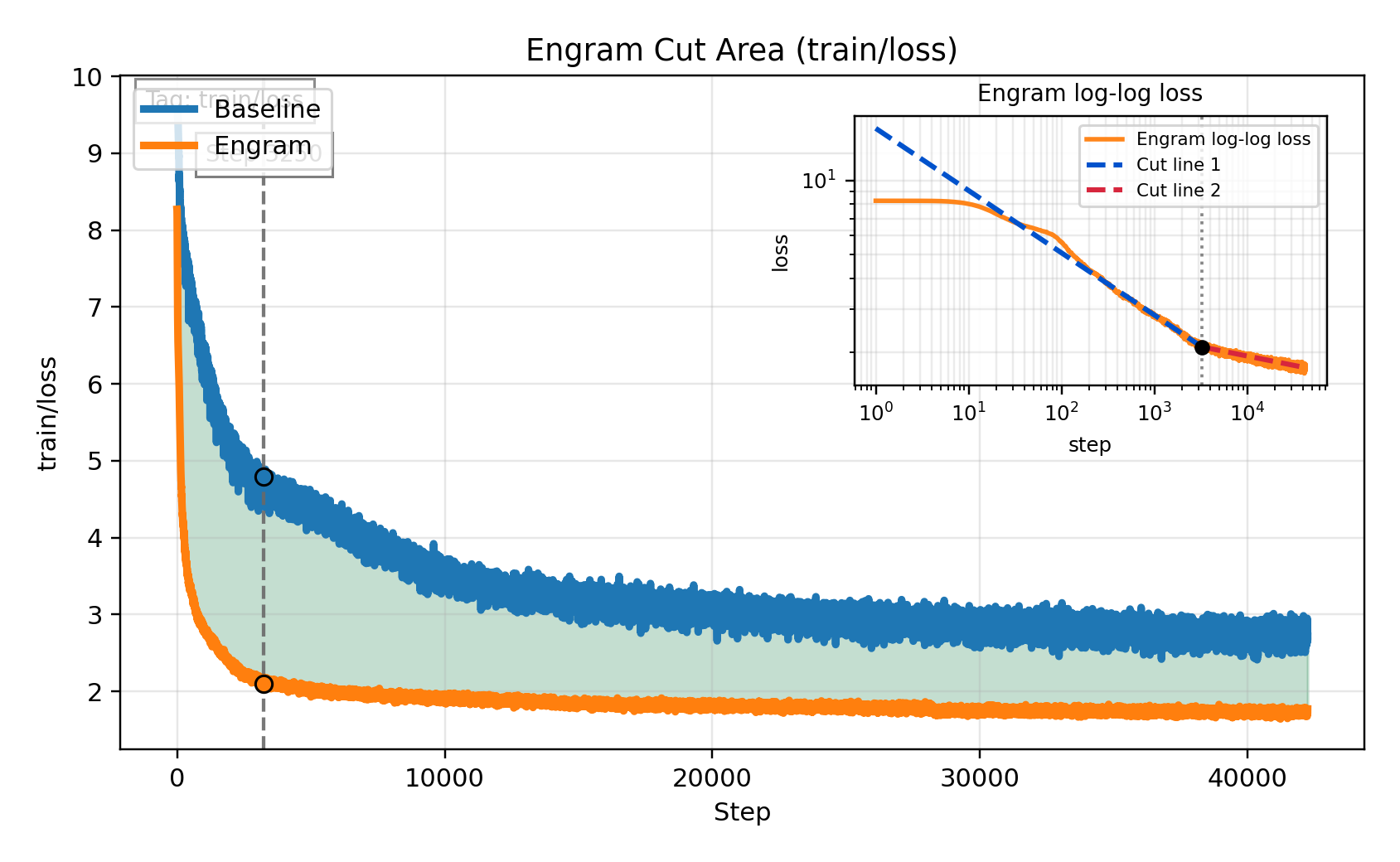}
    \caption{Phase transition in the training loss graph of Toba Trilingual Engram. Inset: Toba Trilingual Engram loss graph on a log-log scale.}
    \label{fig:figure4}
\end{figure*}

The visual representation in Figure \ref{fig:figure4} (Engram Cut Area) quantifies the magnitude of computational resource savings through the green shaded area between the two curves. This area reflects a step efficiency of 80\% in achieving the same accuracy target as the baseline model. The log-log loss inset graph in Figure \ref{fig:figure4} reveals the loss decline dynamics following a power law through two distinct decay phases. Cut Line 1, characterized by a steep slope on the first segment, reflects a highly aggressive learning rate. The Engram module, with a learning rate five times higher, successfully triggers instantaneous statistical memory population, causing the loss value to decrease more rapidly than in the standard transformer architecture. Cut Line 2 represents the transition to a more gradual slope after the transition point, indicating a shift in the model's workload toward long-range context comprehension (global dependencies). Although the rate of decline decelerates, the loss trajectory remains at a level far below the baseline, demonstrating that the Engram memory continues to function as a stable information anchor for the 36 transformer blocks. Between Cut Line 1 and Cut Line 2 lies the transition point (dashed vertical line), the intersection of which is marked by a black dot indicating the saturation point of local statistical memory. At this stage, the 500,000-entry embedding table has internalized trilingual syllable patterns, signifying that the model is ready to transition to a more complex optimization phase.

\subsection{Engram Gradient Observation}
The analysis of numerical stability and learning intensity of the memory module is reflected through the fluctuations of the Engram gradient norm during the optimization process. The dynamics recorded during the early phase provide empirical evidence regarding the transition from passive initialization to the active role of memory in accelerating overall model convergence.

\begin{figure*}[ht]
    \centering
    \includegraphics[width=\textwidth]{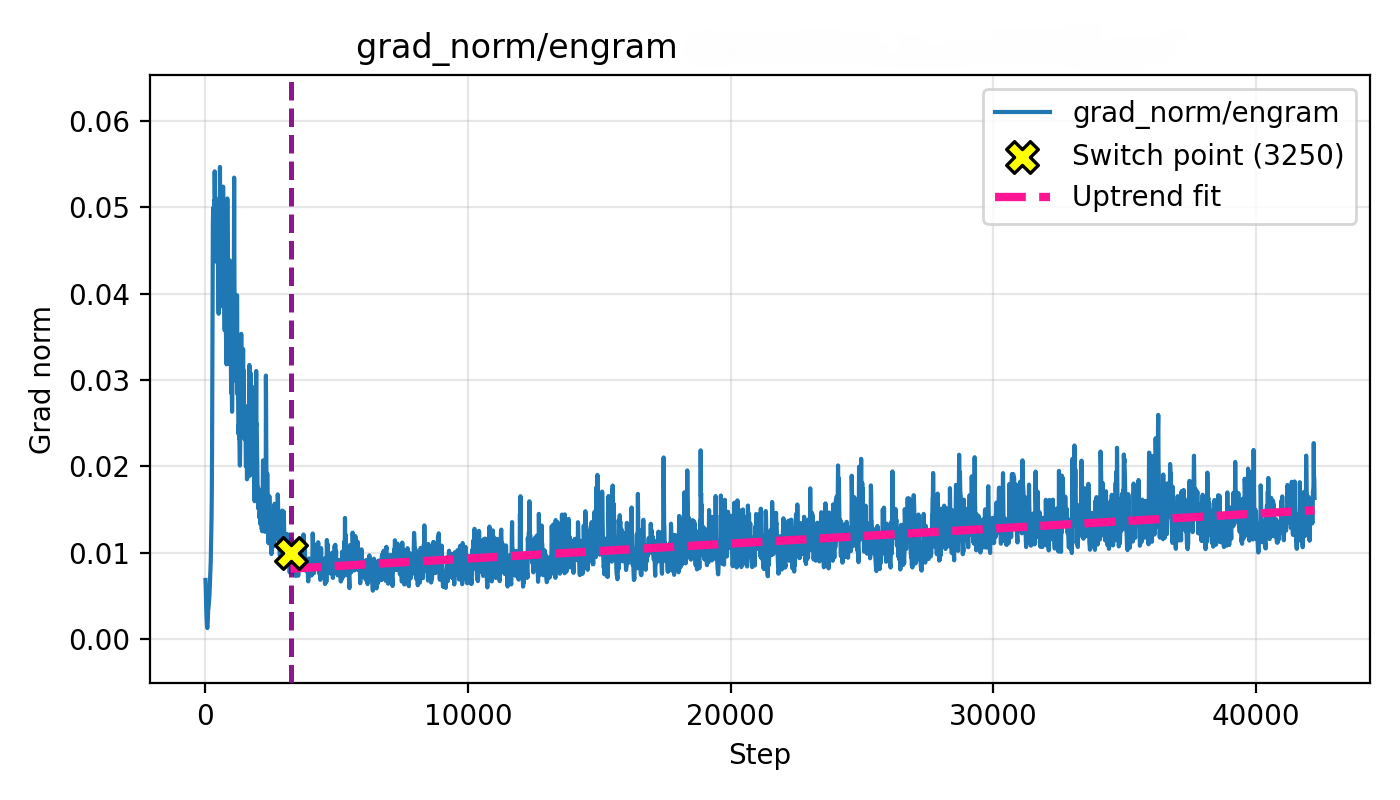}
    \caption{Engram Layer Gradient Norm Graph.}
    \label{fig:figure5}
\end{figure*}

Observation of the gradient dynamics in the Engram module shown in Figure \ref{fig:figure5} reveals a characteristic initialization phenomenon during the early training phase. At this stage, the Engram memory representations do not yet contain meaningful contextual information, as indicated by low gradient norm values. This condition is attributable to the zero-value initialization of the Engram memory embeddings. This dynamic changes abruptly, reaching a switch point at step 3,250, where a gradient intensity overshoot occurs, surpassing the 0.05 threshold. The sharp increase followed by a decline in gradient norm marks an active identification phase in which the model massively adjusts the weights of the Engram embedding table to capture fundamental morphological features from the trilingual corpus.

The significant weight changes in the Engram layer correlate with the vertical loss decline observed in Figure \ref{fig:figure4} during the early training phase. The functional synergy between the adaptive memory module and the transformer components enables the model to map n-gram statistics into the latent space instantaneously, producing a convergence trajectory that is far steeper than that of the baseline architecture. This sharp decline demonstrates that the memory system successfully assumes the burden of surface-level text pattern mapping, freeing the 1.2B parameter capacity of the backbone to focus on optimizing deeper linguistic dependencies.

This drastic loss decline is a manifestation of the emergence of induction heads—internal circuit mechanisms that enable the model to accurately recognize and predict repetitive patterns within a given context. The gradual upward trend (uptrend fit) of the gradient norm after passing the switch point marks the maturation phase of these circuits, which serves to strengthen the synchronization between statistical memory and hidden states processed by the Transformer layers. This induction mechanism plays a critically important role in modeling Austronesian languages, which are characterized by high degrees of morphological agglutination. The availability of local statistical memory enables the model to perform dynamic copying and generalize morphological patterns more efficiently, thereby accelerating the stabilization of model performance. The cumulative effect of this mechanism contributes to the achievement of stable convergence with a final loss value of 1.7996.

\section{Discussions}
\subsection{Implications of the Transformer-Engram Architecture on Computational Efficiency}
The integration of the Engram layer fundamentally reconstructs the information processing hierarchy within the Toba Trilingual architecture by delegating the internalization of local lexical dependencies to a dedicated memory module before data reaches the global attention layers. The placement of this module after the third block enables the model to immediately identify inter-token correlations at the syllable level through a multiplicative-XOR hashing mechanism that maps 2-gram and 3-gram windows instantaneously. The use of the Engram layer reinforces the efficiency of the Toba tokenization output, which is inherently designed to capture agglutinative linguistic patterns in Batak and Minangkabau. By offloading the burden of redundant surface-level text pattern searching to the memory module, the primary self-attention circuits allocate their full computational capacity to modeling long-range semantic dependencies as well as more complex syntactic structures and sentence logic.

The presence of the adaptive Engram memory system systematically contracts the model's search space by providing statistical "anchors" that constrain the probability distribution during the token prediction stage. By leveraging a dedicated embedding table of 500,000 $\times$ 768 dimensions, the Engram module provides precise initial estimates based on the occurrence frequencies of n-gram patterns internalized during the optimization process. The adaptive gating mechanism, which implements RMSNorm and Scaled Dot Product, ensures that only contextually relevant memory signals are injected into the main transformer pipeline. This search space contraction yields a training step efficiency of 80\%, enabling the model to reach the optimal convergence point of 1.7996 in only 12,973 steps. The achievement of the target loss within a total relative time of 23.38 hours confirms that the use of external memory, initiated through the gradient surge at the switch point at step 3,250, drastically reduces the iterations required for the model to comprehend regional language structures.

\subsection{Analysis of the Engram Strategy}
The implementation of the Engram strategy dramatically reduces computational overhead through a reduction in training steps of over 400\% compared to the standard transformer baseline scenario. The green shaded area in Figure \ref{fig:figure4} (Engram Cut Area) indicates the magnitude of processing power savings (floating point operations) achieved, where the model was able to reach the target loss value in only 12,973 steps, while the conventional architecture required more than 70,000 steps to approach a comparable level of convergence. This massive efficiency gain has a direct impact on optimizing server infrastructure utilization, as GPU operational duration can be reduced, which technically minimizes the risk of performance degradation due to thermal throttling and significantly lowers operational energy costs. The success of this computational overhead reduction affirms that the addition of parameters through a 500,000-entry Engram table constitutes a high-return investment for regional language models. Although the integration of the Engram table increases the total parameter capacity to 1.2B parameters, the use of hashing mechanisms and adaptive gating ensures that the per-step computational cost remains low, as it does not involve intensive matrix multiplication operations repeated at every layer. This strategy enables the training of a 36-block deep model to run optimally on hardware with limited VRAM, demonstrating that intervention at the statistical memory level is far more effective in accelerating convergence than merely increasing the number of backbone parameters linearly.

\section{Conclusion}
The implementation of the Engram layer successfully localizes linguistic dependencies at the syllable level, which directly triggers the early formation of induction head mechanisms. This phenomenon accelerates the model's ability to recognize repetitive language patterns, resulting in a convergence trajectory that is far steeper than that of conventional transformer architectures. This acceleration is confirmed through the achievement of a loss value of 1.7996 in a significantly shorter timeframe, demonstrating that the synergy between the Engram module and the transformer backbone is capable of optimizing latent space mapping instantaneously from the initialization stage.

The integration of this memory system carries strategic consequences in the form of enhanced training efficiency for agglutinative languages, particularly regional languages in Indonesia such as Batak and Minangkabau. The Engram module's capacity to capture morphological features through 2-gram and 3-gram units effectively narrows the model's search space, enabling the reasoning process within internal layers to become more focused. The ultimate impact is a massive reduction in computational resources, reflected in an 80\% reduction in training steps that enables a model with 1.1B parameter capacity to achieve optimal convergence. This success affirms that convergence optimization based on statistical memory constitutes a highly feasible solution for the development of regional language models under constraints of limited data and computational infrastructure.

This approach broadly offers opportunities and potential for the preservation of other regional languages sharing similar characteristics (syllabic agglutination), which are overwhelmingly constrained by the limited availability of high-quality datasets, through the tokenization initiated by TOBA-LM. This extends the capability impact of advanced Generative AI applications for language preservation across the Austronesian region.

\section*{Acknowledgement}

The author thank Institut Teknologi Del, fellows in Bandung Fe Institute, and colleagues in Ina-AI Co. for support. All faults remain authors'.

\end{document}